\documentclass[conference,10pt]{IEEEtran}
\usepackage{bbm}
\usepackage{amsmath}
\usepackage{acronym}  
\usepackage[dvips]{color}
\usepackage{epsf}
\usepackage{times}
\usepackage{epsfig}
\usepackage{notoccite}
\usepackage{graphicx}
\usepackage{epstopdf}
\usepackage{pstricks}
\usepackage{amssymb}
\usepackage{amsxtra}
\usepackage{here}
\usepackage{rawfonts}
\usepackage{float}
\usepackage{times}
\usepackage{url}
\usepackage{cite}
\usepackage{caption}
\usepackage{subcaption}
\usepackage{algorithm}
\usepackage{algpseudocode}
\usepackage{blindtext}
\usepackage{enumitem}
\usepackage{xcolor,cite,etoolbox}
\usepackage{relsize}
\usepackage{lipsum}
\usepackage{graphicx}
\usepackage{tabularx}
\usepackage{xparse}
\usepackage{array}
\newcolumntype{P}[1]{>{\centering\arraybackslash}p{#1}}
\usepackage{mleftright}

\usepackage{mathtools}

\usepackage{graphics}
\usepackage{physics}
\usepackage{amssymb}
\usepackage{siunitx}
\include{newcommands}
\usepackage{multicol}

\usepackage[nomain,acronym,shortcuts]{glossaries}
\makeglossaries
\newcommand*{\acro}[3][]{\newacronym[#1]{#2}{#2}{#3}}

\acro{D2D}{device-to-device}
\acro{SIR}{signal-to-interference-ratio}
\acro{SINR}{signal-to-interference-plus-noise-ratio}
\acro{PCP}{Poisson cluster process}
\acro{CoMP}{coordinated multi-point}
\acro{BS}{base station} 
\acro{MD-CoMP}{macrodiversity CoMP transmission}
\acro{MAC}{medium-access-control}
\acro{JT-CoMP}{joint transmission CoMP}
\acro{CoMP-JT}{coordinated multipoint joint transmission}
\acro{SBS}{small base station}
\acro{MDSD}{multiple devices to a single device}
\acro{MDS}{maximum distance separable}
\acro{SCN}{small cell network}
\acro{PPP}{Poisson point process}
\acro{TCP}{Thomas cluster process}
\acro{CSI}{channel state information}
\acro{PDF}{probability distribution function}
\acro{PMF}{probability mass function}
\acro{RV}{random variable}
\acro{i.i.d.}{independently and identically distributed}
\acro{MBMS}{multimedia broadcasting multicasting service}
\acro{EE}{energy efficiency}
\acro{HCP}{hard-core placement}
\acro{CCDF}{complementary cumulative distribution function}
\acro{CDF}{cumulative distribution function}
\acro{PC}{probabilistic caching}
\acro{RC}{random caching}
\acro{CPF}{caching popular files} 
\acro{PGFL}{probability generating functional}
\acro{KKT}{Karush-Kuhn-Tucker}
\acro{PGF}{point generating function}
\acro{SCA}{successive convex approximation}
\acro{HD}{high-definition}
\acro{FHD}{full-high-definition}
\acro{UHD}{ultra-high-definition}
\acro{VR}{virtual reality}
\acro{AR}{augmented reality}
\acro{5G}{fifth-generation}
\acro{QoS}{quality-of-service}
\acro{QoE}{quality-of-experience}
\acro{IoT}{Internet of Things}
\acro{MHCPP}{Matern hardcore point process}
\acro{LoS}{line-of-sight}
\acro{NLoS}{non-line-of-sight}
\acro{PSD}{power spectral density}
\acro{MEC}{mobile edge computing}
\acro{E2E}{end-to-end}
\acro{THz}{terahertz}
\acro{CLT}{central limit theorem}
\acro{HQ}{High Quality}
\acro{eMBB}{enhanced mobile broadband}
\acro{URLLC}{ultra reliable low latency communications}
\acro{mmWave}{millimeter wave}
\acro{EVT}{extreme value theory}
\acro{GEV}{generalized extreme value}
\acro{LIS}{large intelligent surface}
\acro{RIS}{reconfigurable intelligent surface}
\acro{RF}{radio frequency}
\acro{UE}{user equipment}
\acro{MIMO}{multiple-input multiple-output}
\acro{EVaR}{entropic value-at-risk}
\acro{DNN}{deep neural network}
\acro{MDP}{Markov decision process}
\acro{RL}{reinforcement learning}
\acro{RNN}{recurrent neural network}
\acro{ANN}{artificial neural networks}
\acro{LSTM}{long short-term memory}
\acro{ReLu}{rectified linear unit}
\acro{VaR}{value-at-risk}
\acro{SNR}{signal-to-noise ratio}
\acro{AoSA}{array of subarray}
\acro{XR}{extended reality}
\acro{AoA}{angle of arrival}
\acro{ULA}{uniform linear array}
\acro{AoD}{angle of departure}
\acro{EM}{electromagnetic}
\acro{HRLLC}{s high-rate and high-reliability low latency communications}
\acro{6DoF}{six degrees of freedom}
\acro{MR}{mixed reality}
\acro{PAPR}{peak to average power ratio}
\acro{OFDM}{orthogonal frequency-division multiplexing}
\acro{OFDMA}{orthogonal frequency-division multiple access}
\acro{SC-FDM}{single carrier frequency-division multiplexing}
\acro{ToA}{time of arrival}
\acro{MUSIC}{multiple signal classification}
\acro{IoE}{Internet of Everything}
\acro{DT}{digital twin}
\acro{PT}{physical twin}
\acro{CT}{cyber twin}
\acro{DRL}{deep reinforcement learning}
\acro{FL}{federated learning}
\acro{DL}{deep learning}
\acro{CRAS}{connected robotics and autonomous system}
\acro{CL}{continual learning}
\acro{MSE}{mean squared error}
\acro{EWC}{elastic weight consolidation}
\acro{ML}{machine learning}
\acro{GD}{gradient descent}
\acro{MLP}{multi layer perceptron}
\acro{TL}{transfer learning}
\acro{AI}{artificial intelligence}

\usepackage{datetime}
\usepackage{amssymb}
\usepackage{subcaption}
\usepackage{verbatim}

\begin{document}

\title{Edge Continual Learning for Dynamic Digital Twins over Wireless Networks\vspace{-0.3cm}}
\author{Omar Hashash, Christina Chaccour, and Walid Saad \\  Wireless@VT, Bradley Department of Electrical and Computer Engineering, Virginia Tech, Arlington, VA, USA.\\ Email:\{omarnh, christinac, walids\}@vt.edu}
\maketitle
\begin{abstract}
Digital twins (DTs) constitute a critical link between the real-world and the metaverse. To guarantee a robust connection between these two worlds, DTs should maintain \emph{accurate} representations of the physical applications, while preserving \emph{synchronization} between real and digital entities.
In this paper, a novel edge continual learning framework is proposed to accurately model the evolving affinity between a physical twin (PT) and its corresponding cyber twin (CT) while maintaining their utmost synchronization.
In particular, a CT is simulated as a deep neural network (DNN) at the wireless network edge to model an autonomous vehicle traversing an episodically dynamic environment. As the vehicular PT updates its driving policy in each episode, the CT is required to concurrently adapt its DNN model to the PT, which gives rise to a de-synchronization gap. Considering the \emph{history-aware} nature of DTs, the model update process is posed a dual objective optimization problem whose goal is to jointly minimize the loss function over all encountered episodes and the corresponding de-synchronization time. As the de-synchronization time continues to increase over sequential episodes, an elastic weight consolidation (EWC) technique that regularizes the DT history is proposed to limit de-synchronization time. Furthermore, to address the \emph{plasticity-stability} tradeoff accompanying the progressive growth of the EWC regularization terms, a modified EWC method that considers fair execution between the historical episodes of the DTs is adopted. Ultimately, the proposed framework achieves a simultaneously accurate and synchronous CT model that is robust to \emph{catastrophic forgetting}. Simulation results show that the proposed solution can achieve an accuracy of $90\%$ while guaranteeing a minimal de-synchronization time.
\end{abstract}
\vspace{-0.2cm}
\section{Introduction}
\indent \Acp{DT} are central to the digital transformation that we are currently witnessing. In essence, \acp{DT} will be a driving force for an end-to-end digitization beyond the Industry 4.0 borders \cite{khan2022digital}. 
By extending the fundamental \ac{DT} concept towards massive twinning, a plethora of complex real-world \acp{DT} are expected to emerge~\cite{han2021multi}. This is crucial for enabling anticipated \ac{IoE} applications such as extended reality and connected robotics and autonomous
systems~\cite{chaccour2022seven}. 
Guaranteeing a high-fidelity representation of such \acp{DT} that have a dynamic nature constrains the underlying communication network with a set of stringent wireless requirements.
For instance, extreme reliability, hyper-mobile connectivity, near-zero latency, and high data rates are necessary to sustain high \ac{QoS} for such dynamic \acp{DT}. One measure that can bring us closer to guaranteeing such requirements is the integration of \acp{DT} with an intelligent edge backbone.
Furthermore, an \ac{AI}-native edge plays a pivotal role in supporting fully-autonomous \ac{IoE} services in 6G networks \cite{chaccour2021edge}.
In fact, deploying \acp{DT} for autonomous IoE applications operating in non-stationary real-world scenarios necessitates adopting an intelligent wireless edge that can seamlessly adapt the digital duplicate to the changes in the underlying application states. \\
\indent Several works \cite{lu2021adaptive,lin2021stochastic,lu2020low,sun2020dynamic} studied the implementation of \acp{DT} at the edge in an attempt to meet their stringent requirements and meet their challenges.
The authors in \cite{lu2021adaptive} proposed coupling \ac{DRL} with \ac{TL} to solve the \ac{DT}-edge placement and migration problems in dynamic environments. The work in \cite{lin2021stochastic} proposed a dynamic congestion control scheme to manage the stochastic demand in \ac{DT}-edge networks. The authors in \cite{lu2020low} developed a solution for the dynamic association problem in DT-edge networks while balancing the accuracy and cost of a blockchain-\ac{FL} scheme. In~\cite{sun2020dynamic}, the authors studied the use of a \ac{DT} for capturing the air-ground network dynamics that assist network elements in designing incentives for a \ac{DT}-edge \ac{FL} scheme. While the works in \cite{lu2020low,sun2020dynamic,lu2021adaptive,lin2021stochastic} shed light on interesting issues that accompany \ac{DT}-edge networks from different dynamic perspectives, they consider major limiting assumptions.
The works in \cite{lu2020low,sun2020dynamic,lin2021stochastic} assume the \acp{DT} to operate under ideal stationary conditions. Nonetheless, realistic \acp{DT} that emulate a digital replica of novel \ac{IoE} applications operate in dynamic non-stationary environments. Meanwhile, even though the work in \cite{lu2021adaptive} considers non-stationary conditions when using \ac{TL}, it disregards the knowledge gained with the \emph{continuous} evolution of the \acp{DT}. 
Clearly, there is a lack in works that comprehensively consider the practical operation of evolving \acp{DT} under real dynamic considerations.  

In fact, enabling a dynamic \ac{DT} in a massive twinning context is governed by defining the duality between physical and digital entities.
This duality is triggered on multiple fronts that require merging new dimensions and metrics, including: 
(i) \textit{synchronization} between the \ac{PT} and the \ac{CT} to preserve the real-time interaction and diminish the possibility of interrupting the \ac{CT} operations and simulations, 
(ii) \textit{accuracy} of the \ac{CT} in reflecting the real-time PT status by maintaining a precise duplicate throughout the various phases experienced by the PT, and 
(iii) \textit{history-awareness} of the evolving \ac{DT} states by incorporating the knowledge attained from past experiences into future states.
\emph{Hence, a fundamental step towards achieving a massive twinning framework can be accomplished by enabling simultaneously synchronous, accurate, and history-aware \acp{DT} that can continuously operate in dynamic non-stationary environments.}

The main contribution of this paper is a novel edge \ac{CL} approach to accurately model the evolving affinity between a \ac{PT} and its corresponding \ac{CT} in a real dynamic environment, while preserving their synchronization.
In particular, a CT is simulated as a \ac{DNN} at the wireless network edge to model an autonomous vehicle traversing through an episodically dynamic environment. 
This vehicular PT is equipped with \ac{IoT} sensors that capture the relevant data used in determining the autonomous driving actions. As the distribution of the \ac{IoT} data changes with each episode, the vehicular agent adapts its driving policy accordingly.
Consequently, the \ac{CT} must update its model to preserve the \ac{DT} accuracy, which leads to a de-synchronization gap between the twins.
While taking the history-awareness of \acp{DT} into account, the \ac{CT} model update process is formulated as a dual objective optimization problem to minimize the loss over all encountered episodes while reducing de-synchronization time. 
To limit the progressive increase in de-synchronization time over episodes, we propose a \ac{CL}-based \ac{EWC} technique that contracts the ongoing historical experience as regularization terms into the learning cost. 
As the impact of \ac{EWC} regularization on \ac{CT} evolution increases with number of episodes encountered, we use a variant of \ac{EWC} to address the \emph{plasticity-stability} effect in \acp{DT}. \textit{To the best of our knowledge, this is the first work to model \acp{DT} in a dynamic environment by addressing the synergy between synchronization, accuracy, and ongoing history.} Simulation results show that our proposed method to model \acp{DT} can outperform state-of-art benchmarks by providing near optimal accuracy measures over all encountered episodes within the minimal de-synchronization time.
\vspace{-0.15cm}
\section{System Model}
\subsection{DT-Edge Model}
\begin{figure}
	\includegraphics[scale=0.39]{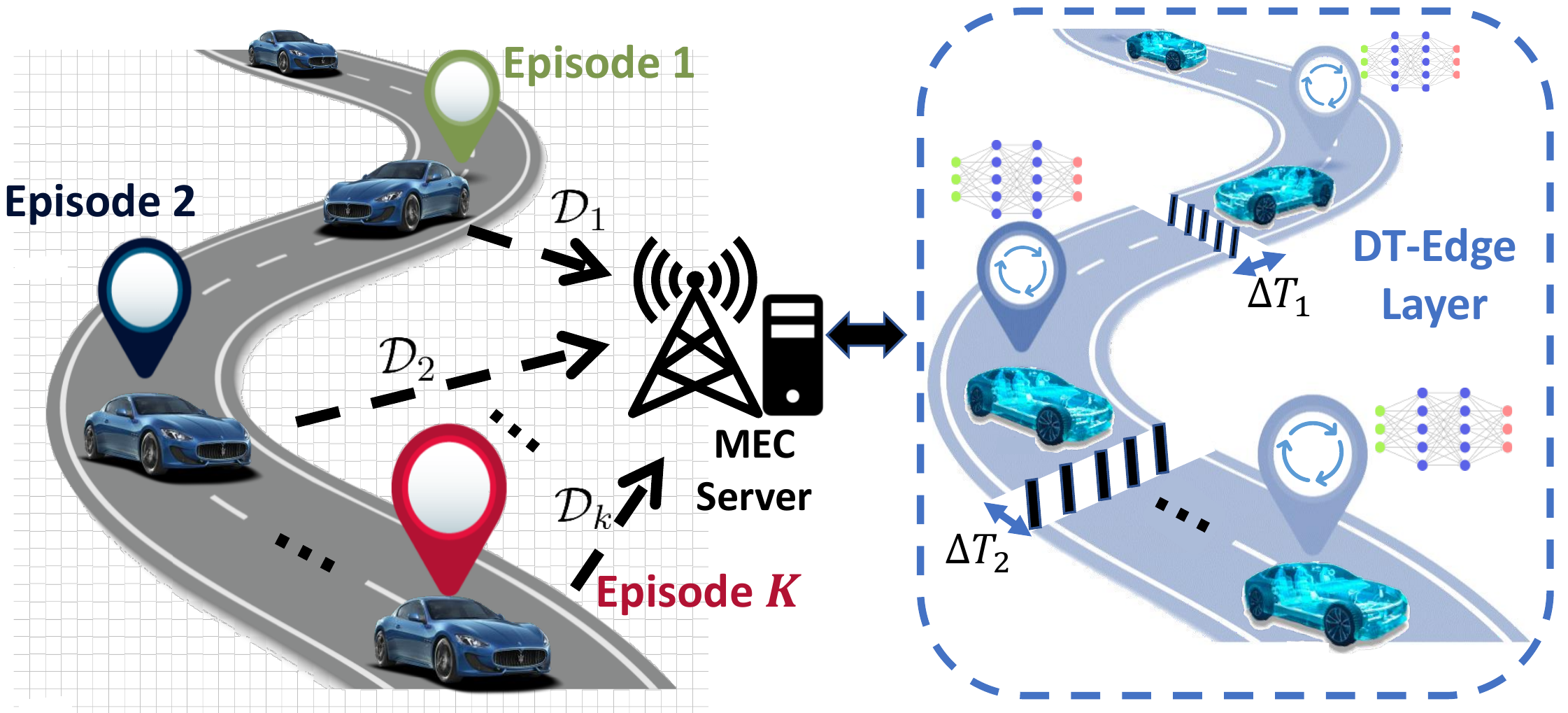}
	\caption{\small{Illustration of the edge dynamic digital twins system performing a simulation of the physical system in a non-stationary environment}}
	\label{System Model}
	\vspace{-0.50cm}
\end{figure}

Consider a \ac{DT} system in which the \ac{PT} is an autonomous vehicle connected to a \ac{BS}, as shown in Fig.~\ref{System Model}. To empower this vehicle with autonomous functionality, it is equipped with a large set of \ac{IoT} sensors that continuously capture data used in executing real-time driving decisions. 
For instance, the type of data generated by the \ac{IoT} sensors can range from intra-vehicular analytics to inter-vehicular range sensing that are necessary to fine-tune the vehicle's speed. Moreover, the \ac{BS} is equipped with a \ac{MEC} server that runs a real-time \ac{DT} simulation of the autonomous driving vehicle. In particular, the vehicular \ac{CT} simultaneously replicates the \ac{PT}'s driving decisions as a function of the \ac{IoT} data uploaded to the \ac{DT}-edge layer. This real-time \ac{DT} simulation is managed in a deep learning framework, where a \ac{DNN} model $\mathcal{M}_0$ that represents the \ac{CT} is trained over a dataset $\mathcal{D}_0$ comprising historical actions exhibited by the \ac{PT}. The upload time of \ac{IoT} data to the \ac{DT}-edge layer is assumed to be negligible.

As a result of the dynamic changes in the space, time, as well as the driving conditions; the environment experienced by the vehicle will vary in a \emph{non-stationary} fashion. These dynamic and non-stationary variations are reflected in the generated \ac{IoT} data readings and their respective probability distributions. Thus, the \ac{PT} and its respective \ac{CT} are required to adjust their driving policy to cope with those variations. 

Henceforth, we consider a \ac{PT} to be operating in a dynamic environment, where the environment changes over a set $\mathcal{K}$ of $K$ sequential time episodes~\cite{sun2022learning}. For tractability, although the inter-episode variations are non-stationary, we assume that the intra-episode environment variations are stationary. In our model, the learning agent has full knowledge of the time frames, i.e., the start and end of each episode. As a result of dynamic changes in the \ac{PT} environment, the learning agent of the \ac{PT} autonomously updates its driving policy.
\emph{Consequently, considering the history-aware nature of \acp{DT}, the \ac{CT} should adapt its \ac{DNN} model} \emph{to encompass the current episode $k \in \mathcal{K}$ while reflecting the knowledge from its past experiences.} 
  
At the start of each episode $k$, a dataset $\mathcal{D}_k = \{\boldsymbol{x}_{ki},y_{ki}\}_{i=1}^{i=I_k}$ of the vehicle actions is collected, where $I_k = |\mathcal{D}_k|$ and the pair $\{\boldsymbol{x}_{ki},y_{ki}\} \sim \mathcal{P}_k$ represents the $i^{\mathrm{th}}$ data vector of the IoT sensor measurements and the corresponding decision taken by the vehicle, during episode $k$, respectively. Then, this dataset $\mathcal{D}_k$ is uploaded to the DT-edge layer to update a DNN model $\mathcal{M}_k$ in episode $k$.
\vspace{-0.20cm}
\subsection{Learning over the DT-Edge}
The model $\mathcal{M}_k$ is assumed to have $l$ fully-connected layers where $l = 1,2,\ldots,L$. In addition, each layer is composed of $a_{lm}$ neurons, where $m=m_1,m_2,\ldots,m_l$ is the number of neurons in each layer. Moreover, $\mathcal{M}_k$ is characterized with weights $\boldsymbol{w}_k^{n_k} \in \mathbb{R}^{n}$ reached upon performing $n_k$ training iterations, with $n = \sum_{l=1}^{L-1} m_l m_{l+1}$ being the dimension size of the model weights vector.
Due to the aforementioned dynamic spatio-temporal conditions, the data probability distribution $\mathcal{P}_k$, where $\mathcal{D}_k \sim \mathcal{P}_k$, experiences non-stationary variations with respect to $k$ episodes. 
Hence, to procure a comprehensive update that describes the entire \ac{PT} evolution, the \ac{CT} should adapt its model to the current episode $k$ while reflecting the experience gained over the preceding $k-1$ episodes.
In accordance, we define a loss function that measures the performance of $\mathcal{M}_k$ over all collected \ac{PT} actions, as follows:
\vspace{-0.28cm}
\begin{equation}
\mathcal{L}(\boldsymbol{w}_k^{n_k}, \mathcal{A}_k) = \frac{1}{|\mathcal{A}_k|} \sum_{j=0}^{k}\sum_{i=1}^{I_k} \psi(\boldsymbol{w}_k^{n_k},\boldsymbol{x}_{ji},y_{ji}),
\vspace{-0.28cm}
\end{equation}
where  $\mathcal{A}_k = \bigcup_{j=0}^{k} \mathcal{D}_j$ is the \emph{non-iid} accumulated dataset of all \ac{PT} actions until current episode $k$ and $\psi(\cdot)$ is a generic loss function.
The training process over the \emph{non-iid} dataset is then carried out using \ac{GD} iterations to minimize the loss function as follows: $ \boldsymbol{w}_{k}^{n_k} \leftarrow \boldsymbol{w}_{k}^{n_k-1} - \eta \nabla \mathcal{L}(\boldsymbol{w}_k^{n_k-1}, \mathcal{A}_k), $
where $\eta$ is the learning rate.
Since the \ac{DT} simulation is an explicitly synchronous process, the \ac{CT} cannot perform simulation and learning tasks simultaneously. Thus, we assume that the simulation process is halted throughout the updating process of $\mathcal{M}_k$, which gives rise to a synchronization gap between the twins. 
To capture the time during which the \ac{DT} is out-of-service, we define the \emph{de-synchronization time} as the time needed to update $\mathcal{M}_k$ during each episode $k$, as follows: 
\vspace{-0.27cm}
\begin{equation}
    \Delta T_k (n_k)  =  \frac{|\mathcal{A}_k|}{f}\Lambda n_k,
    \vspace{-0.1cm}
\end{equation}
where $\Lambda$ is the number of CPU cycles required to train one data sample, and $f$ is the CPU frequency of the \ac{MEC} server. Next, we formulate a dual objective optimization problem to guarantee \emph{episodic history-aware \ac{CT} model updates to accurately represent the \ac{PT}, while limiting the de-synchronization time between the twins.}
\subsection{Problem Formulation}
Our goal is to jointly maximize the \ac{DT}'s accuracy and minimize the corresponding de-synchronization time over each episode. In particular, we tend to guarantee a synchronous \ac{DT} simulation where the \ac{CT} model accurately adapts to each episode sequentially while abiding by the knowledge acquired from previous experience. Thus, it is necessary to find the optimal number of iterations need to learn the corresponding weights while achieving an optimal tradeoff between enhancing model accuracy and limiting de-synchronization time.
Thus, at the beginning of each non-stationary episode $k$ the learning agent needs to optimize the following problem, in a sequential fashion:
\vspace{-0.2cm}
\begin{subequations}
\label{opt1}
\begin{IEEEeqnarray}{s,rCl'rCl'rCl}
& \underset{n_k}{\text{min}} &\quad& \alpha\mathcal{L}(\boldsymbol{w}_k^{n_k}, \mathcal{A}_k) + (1-\alpha) \Delta T_k(n_k),  \label{obj1}\\
&\text{s.t.} &&  \alpha \in [0,1], n_k  \in \mathbb{Z} \label{c3}, 
\vspace{-0.2cm}
\end{IEEEeqnarray}
\end{subequations}
where $\alpha$ is a bias coefficient that controls the accuracy and synchronization tradeoff in our application.\\
\indent This is a dual objective optimization problem that minimizes the loss function and the de-synchronization time to maintain an accurate \ac{CT} model in real-time, by controlling the number of \ac{GD} iterations in each episode.
Solving problem~\eqref{opt1} using classical machine learning schemes is challenging due to the growth of the accumulated dataset size that is included in the \ac{GD} computations over each episode -- a unique feature of \acp{DT} over wireless networks.\\
\indent As the size of the accumulated dataset increases with the number of episodes encountered, the de-synchronization time grows linearly with the additional time required to update the model. Given that the episodes have equal importance and contribution to the learning model, the complete ongoing history of experiences should be considered in the loss function. On the one hand, disregarding the previous \ac{DT} experiences violates the history-awareness requirement of our \ac{DT}. On the other hand, training over the accumulated dataset will jeopardize synchronization and violate \ac{DT} latency requirements. Thus, it is necessary to propose a novel solution that can ensure a swift model adaptation in each episode while fulfilling the synchronization and history-aware \ac{DT} requirements. Next, we develop a novel edge \ac{CL} paradigm that balances the accuracy and de-synchronization time in the \ac{CT} update process while taking the historical experience into consideration.
\vspace{-0.15cm}
\section{Edge CL for Dynamic DTs}
\label{solution}
In this section, we present a \ac{CL}-based solution that can mitigate the progressive increase in de-synchronization time which accompanies the growth of accumulated dataset size. In general, \ac{CL} is recognized as an incremental learning scheme that enables agents to learn individual tasks sequentially~\cite{delange2021continual}, without the need to learn over all of the previously encountered tasks' data.
Accordingly, by considering the \ac{DT} performance over each episode as an individual task, \ac{CL} can promote individual episode learning while preserving the knowledge gained from encountered episodes. Thus, by alleviating the need to learn over the accumulated dataset, the increase in de-synchronization time can diminish noticeably.

With respect to our \acp{DT} system, our solution will embrace the \ac{EWC} technique that was introduced to mitigate the \emph{catastrophic forgetting} phenomenon that captures the affinity of DNNs to forget the weights related to previous tasks upon consecutive model updates~\cite{kirkpatrick2017overcoming}.
In particular, \ac{EWC} enables selective regularization of model weights that are essential to reflect the \ac{CT} experience in each episode. 
Hence, a major portion of the \ac{DT} history is conveyed within the \ac{DNN} weights and contracted as quadratic penalty terms inserted into the loss function. Thus, regularizing the history can limit \emph{catastrophic forgetting} and the increase in de-synchronization time.\\
\indent First, the \ac{DNN} weights are acquired by employing a sequential Bayesian approach. In particular, to estimate the posterior distribution of the \ac{DNN} weights given the accumulated data $\mathcal{A}_k$ and, assuming that the episodes are independent, the log posterior distribution is defined according to the Bayes rule~\cite{kirkpatrick2017overcoming} as: $
  \text{log }p(\boldsymbol{w}_{k}^{n_k}|\mathcal{A}_k) \displaystyle \propto \text{log }p(\mathcal{D}_k|\boldsymbol{w}_k^{n_k} ) + \text{log } p(\boldsymbol{w}_{k}^{n_k}|\mathcal{A}_{k-1}),$
where $\text{log }p(\mathcal{D}_k|\boldsymbol{w}_k^{n_k} )$ is the loss in the current episode $k$ and $\text{log } p(\boldsymbol{w}_{k}^{n_k}|\mathcal{A}_{k-1})$ is the prior probability. 
Thus, training the model $\mathcal{M}_k$ is carried out by a minimization over the negative log posterior function to yield the solution weights $\boldsymbol{w}_k^{*}$ in episode $k$ as:
\vspace{-0.13cm}
\begin{equation}\label{tr1}
\vspace{-0.11cm}
    \boldsymbol{w}_k^{*}= \arg\underset{\boldsymbol{w}_k^{n_k}}{\min} \ \left (-\text{log }p(\boldsymbol{w}_{k}^{n_k}|\mathcal{A}_k) \right),
    \vspace{-0.02cm}
\end{equation}
Here,~\eqref{tr1} represents the \ac{DT} simulation that reflects the \ac{DT}'s state in episode $k$. Furthermore, the ongoing \ac{DT} history is implicitly reflected throughout the prior term that provides a bias of the experience gained over the preceding $k-1$ episodes. In essence, the term  $\text{log }p(\mathcal{D}_k|\boldsymbol{w}_k^{n_k} )$ is related to the CT simulation error while minimizing the loss function with respect to $\mathcal{D}_k$ and $\boldsymbol{w}_k^{n_k}$. The prior term $\text{log } p(\boldsymbol{w}_{k}^{n_k}|\mathcal{A}_{k-1})$ is intractable as shown in ~\cite{chen2018lifelong}, which limits the evaluation of the posterior distribution as a direct term. To evaluate this term, one can formulate its Laplace approximation~\cite{huszar2017quadratic}, and, thus, approximate it through its second order Taylor expansion in the neighborhood of $\boldsymbol{w}_{k-1}^{*}$ as: $\text{log }p(\boldsymbol{w}_k^{n_k}|\mathcal{A}_{k-1})  \approx \frac{1}{2} (\boldsymbol{w}_k^{n_k} - \boldsymbol{w}_{k-1}^{*})^{\top} \boldsymbol{H}(\boldsymbol{w}_{k-1}^{*}) (\boldsymbol{w}_k^{n_k} - \boldsymbol{w}_{k-1}^{*})$, 
where $\boldsymbol{H}(\boldsymbol{w}_{k-1}^{*}) = \frac{\partial^{2}}{\partial \boldsymbol{w}_{k}^2 } \text{log } p(\boldsymbol{w}_k|\mathcal{A}_{k-1})\arrowvert_{\boldsymbol{w}_k = \boldsymbol{w}_{k-1}^{*}}$is the Hessian of the log prior probability function evaluated at $\boldsymbol{w}_{k-1}^{*}$. Moreover, the first and second terms of the Taylor expansion are neglected since they are evaluated near the optimal solution $\boldsymbol{w}_{k-1}^{*}$. Furthermore, the Hessian in this case can be represented in terms of the Fisher information matrix that shows the importance of each weight in this episode.  Without loss of generality, the Fisher information matrix is defined as   
$\boldsymbol{F}_{k} = -\mathbb{E}_{(\boldsymbol{x}_k,y_k) \sim \mathcal{D}_k}(\boldsymbol{H}(\boldsymbol{w}_{k}^{*}))$. 
Assuming that the weight parameters are independent of each other, we can estimate the Fisher information matrix in terms of its diagonal elements while setting the rest to zero~\cite{van2019three}. Hence, after quantifying the terms related to the posterior probability, we define the \emph{recursive loss function} that combines both current and \ac{EWC} losses in episode $k$ as follows:
\vspace{-0.28cm}
\begin{equation}
\label{recursive loss}
    \mathcal{L}(\boldsymbol{w}_k^{n_k})= \mathcal{L}(\boldsymbol{w}_k^{n_k}, \mathcal{D}_k)+ \sum_{j=0}^{k-1}\sum_{d} \frac{\lambda}{2} F_{j,dd}\left(w_{k,d}^{n_k} - w_{j,d}^*\right)^2,
    \vspace{-0.2cm}
\end{equation}
where $\mathcal{L}(\boldsymbol{w}_k^{n_k}, \mathcal{D}_k)$ is the CT simulation error on $\mathcal{D}_k$, $F_{j,dd}$ is the diagonal value of the Fisher information matrix in episode $j$ that relates to the weight $d$, $\lambda$ is a hyperparameter that determines the relative weighting of learning the new episode in comparison to remembering the previous episodes experience, $w_{k,d}$ is the weight in episode $k$, and $w_{j,d}^{*}$ is the optimal solution weight for parameter $d$ in episode $j$.

The \ac{EWC} loss term in \eqref{recursive loss} can substantially increase with the number of episodes encountered, which consequently jeopardizes the model's robustness. Hence, the model's ability to \emph{continuously} update its weights will be dominated by the rigid effect of the regularization terms. Eventually, this will lead to a \emph{stable} \ac{CT} model that is constrained by its history and cannot evolve further. On the other hand, when the effect of regularization is minimal, the model is susceptible to losing its experience which leads to a \emph{plastic} \ac{CT} model that is vulnerable to \emph{catastrophic forgetting}. This phenomenon is known as the \emph{stability-plasticity dilemma}~\cite{parisi2019continual}. To address this dilemma, we must equip the \ac{CT} model with higher degrees of freedom to smoothly address stability and plasticity effects along different episodes. To achieve that, we propose adopting a modified \ac{EWC} version called \ac{EWC}++~\cite{chaudhry2018riemannian}. Here, the Fisher information matrix is estimated through a moving average after calculating the Fisher matrix in each episode $k$. Accordingly, the averaged Fisher elements are estimated at the end of each episode $k$ as:
$ \Tilde{F}_{{k,dd}} = \gamma F_{k,dd} + (1-\gamma) \Tilde{F}_{{(k-1),dd}}$, where $\gamma$ is a hyperparamater.
After setting the value of $\gamma$ and calculating the averaged Fisher elements at each episode, the modified recursive loss function can be expressed as:
\begin{equation*}
\small
    \tilde{\mathcal{L}}(\boldsymbol{w}_k^{n_k})= \mathcal{L}(\boldsymbol{w}_k^{n_k}, \mathcal{D}_k)+ \sum_{d} \frac{\lambda}{2} \Tilde{F}_{{(k-1),dd}}\left(w_{k,d}^{n_k} - w_{(k-1),d}^*\right)^{2}.
    \vspace{-0.25cm}
\end{equation*}
This loss deals with the current and previous episode Fisher values along with the previous weights $\boldsymbol{w}_{k-1}^{*}$ only. 
Essentially, implementing this modified loss function can yield a representative \ac{CT} model that \emph{continually} learns while countering the stability-plasticity effects and mitigating the increase in de-synchronization time. Hence, this modified loss function will replace that of \eqref{obj1} to solve problem \eqref{opt1}. Henceforth, the problem in \eqref{opt1} can be then interpreted as a modified loss minimization problem by controlling $n_k$ that achieves the desired balance between accuracy and de-synchronization of the \ac{DT}s according to $\alpha$. Solving this modified loss minimization problem through \ac{GD} iterations will yield the optimal $n_k^{*} $ and its corresponding $\boldsymbol{w}_{k}^{*}$ for episode $k$.
\section{Simulation Results and Analysis}
\begin{figure}[t!]
		\begin{minipage}{0.49\linewidth}
			\centering
			\includegraphics[scale=0.30]{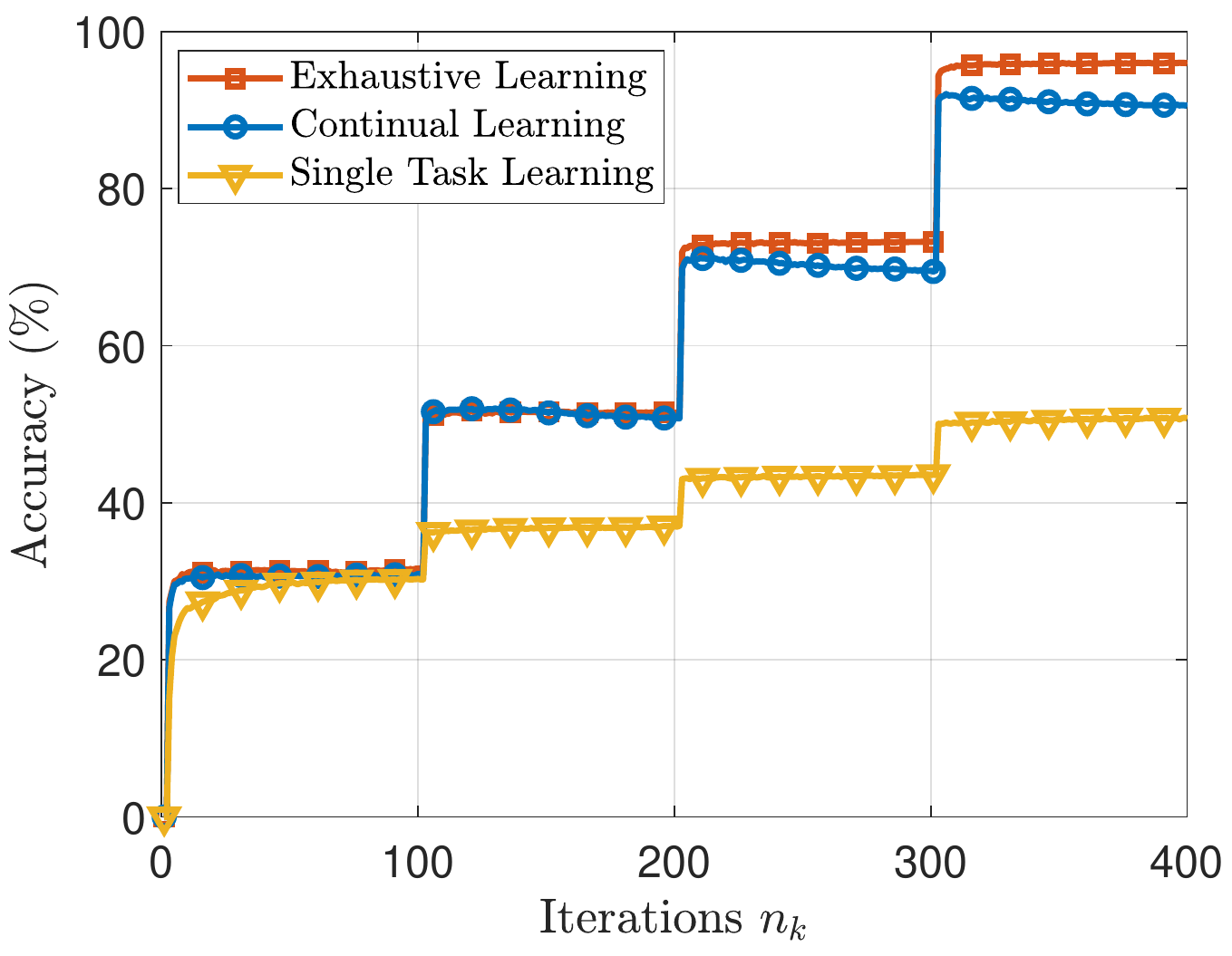}
			\subcaption{}    \label{remembering}
		\end{minipage}
		\begin{minipage}{0.49\linewidth}
			\centering
			\includegraphics[scale=0.3]{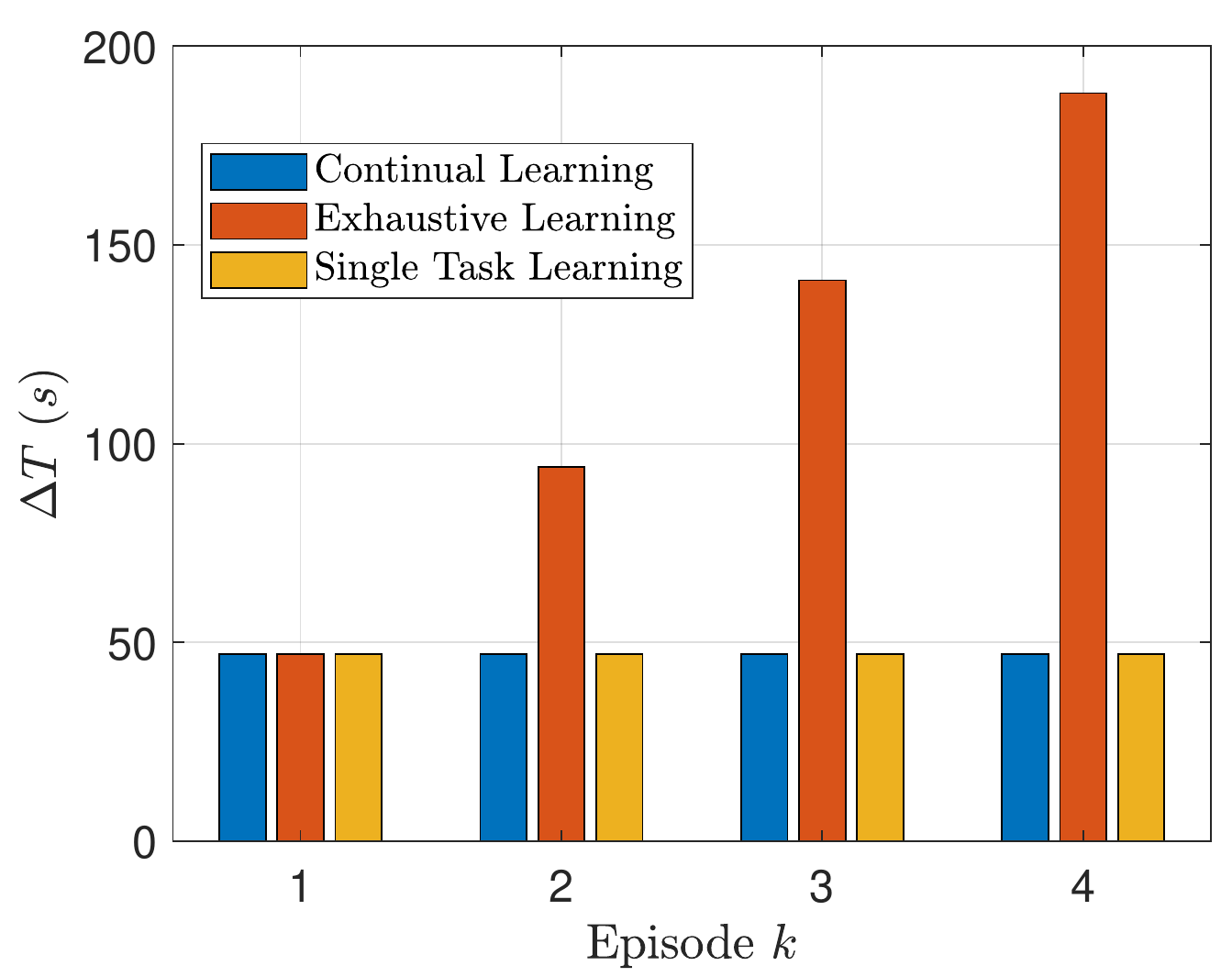}
			\subcaption{}    \label{bar}
		\end{minipage}
		\caption{\small{a) Accuracy (\%) versus the number of iterations $n_k$, b) De-synchronization time upon training over 4 sequential episodes}.}  \label{fig:performance}
		\vspace{-0.3cm}
	\end{figure}

\begin{figure}
	\centering
	\includegraphics[width=0.30\textwidth]{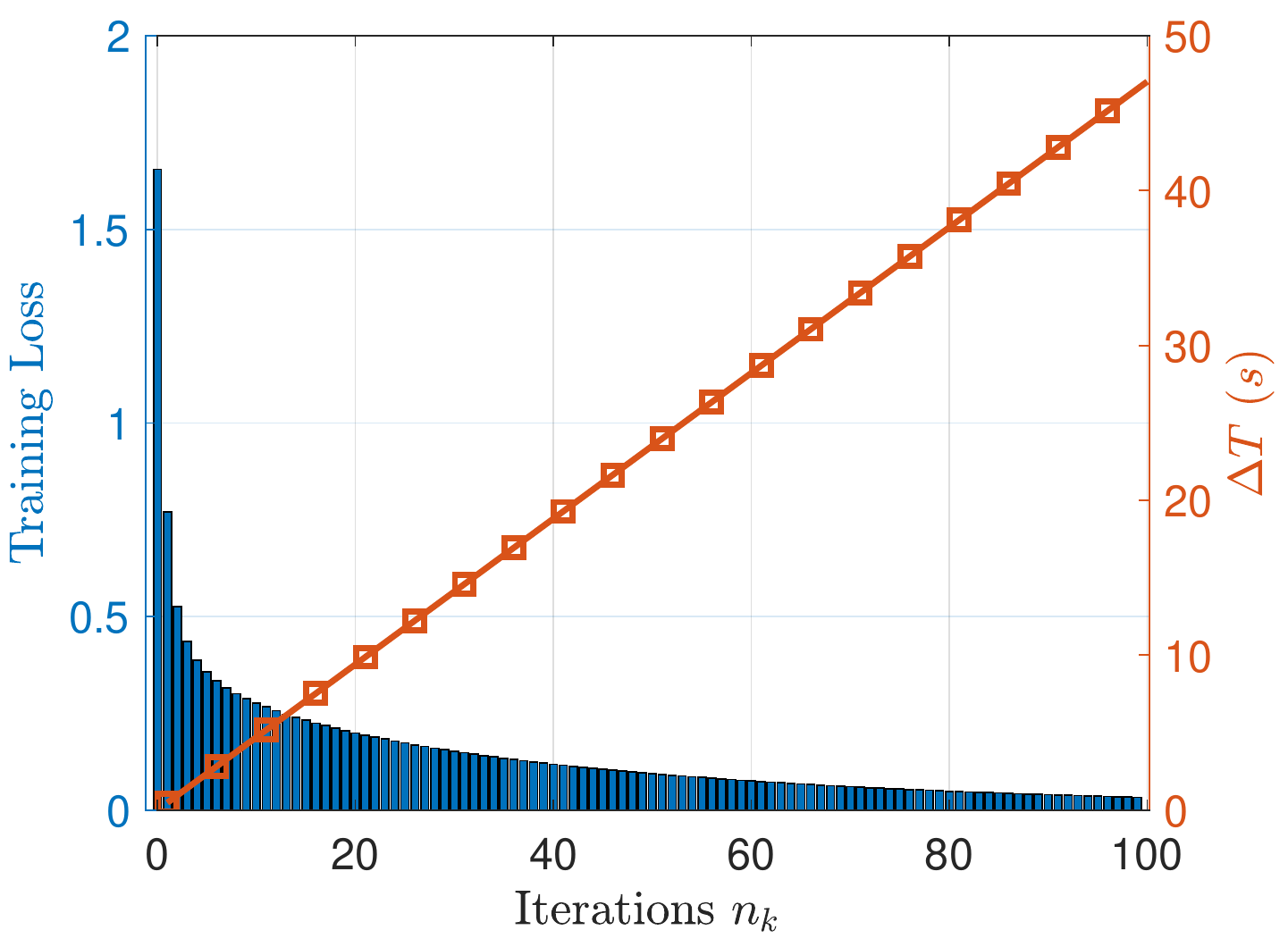}
	\caption{\small{Training loss and de-synchronization time $\Delta T$ (s) versus the number of iterations $n_k$ during one episode.}}
	\label{tradeoff}
	\vspace{-0.5cm}
\end{figure}

For our simulations, we consider a \ac{CT} operating over a \ac{MEC} server at a \ac{BS} for $k = 4 $ sequential episodes. The \ac{CT} must \emph{sequentially} learn a classification task of identifying the handwritten digits from $0$ to $9$ of the permuted MNIST dataset, which is a variant of the standard MNIST dataset having image pixels permuted independently for each episode. 
For our model, we use a \ac{MLP} having $L=2$  hidden layers, each having $m_{l} = 256$ neuron units with ReLu activations $\forall l = \{1,2\}$. We set $\psi(\cdot)$ as the cross-entropy function. Unless otherwise stated, we set $I_k = 15000$, $\eta = 0.01$, $\Lambda= 125440$ cycles/sample, $f= \SI{4}{GHz}$, $\lambda = 75000$, and $\gamma = 0.5$~\cite{chaudhry2018riemannian, zenke2017continual}. A dataset of $1000$ images was procured, by collecting images from the different permutations used to test the trained model. In our experiments, we compare the proposed \ac{CL} solution presented with two benchmark methods: a) An \emph{exhaustive learning} scheme that accumulates all the training data throughout its history and current episode, b) A \emph{single task learning} approach that solely relies on the training data seen in the current episode. \\
\indent Fig.~\ref{remembering} shows the achieved accuracy on the test set over 400 training iterations for the different methods, where each 100 iterations refers to one sequential episode. From Fig.~\ref{remembering}, we observe that the proposed CL solution achieves near optimal performance reaching an accuracy of $90\%$, compared to the optimal solution of $95\%$ resulting from the exhaustive learning scheme. 
Meanwhile, the single task learning method did not retain all of its knowledge between episodes which resulted in an accuracy of only $50\%$.
The results showcase how the proposed \ac{CL} method is capable of learning incrementally without revisiting data from previous episodes.\\
\indent Fig.~\ref{bar} presents the de-synchronization time in each episode. Here, the exhaustive learning scheme achieves a high accuracy, at the expense of a significantly increased de-synchronization time. In this figure, we observe de-synchronization times of $\SI{45}{s}$ and $\SI{190}{s}$ in episodes $1$ and $4$, respectively. In contrast, the proposed \ac{CL} solution has a clear advantage by maintaining its de-synchronization time at $45$ s throughout the $4$ episodes, while reaching a near optimal accuracy. Although single task learning has a similar de-synchronization time to \ac{CL}, this comes at the expense of a deteriorated accuracy of $50\%$ at the end of the 4 episodes.
\begin{figure}[t!]
	\centering
	\includegraphics[width=0.30\textwidth]{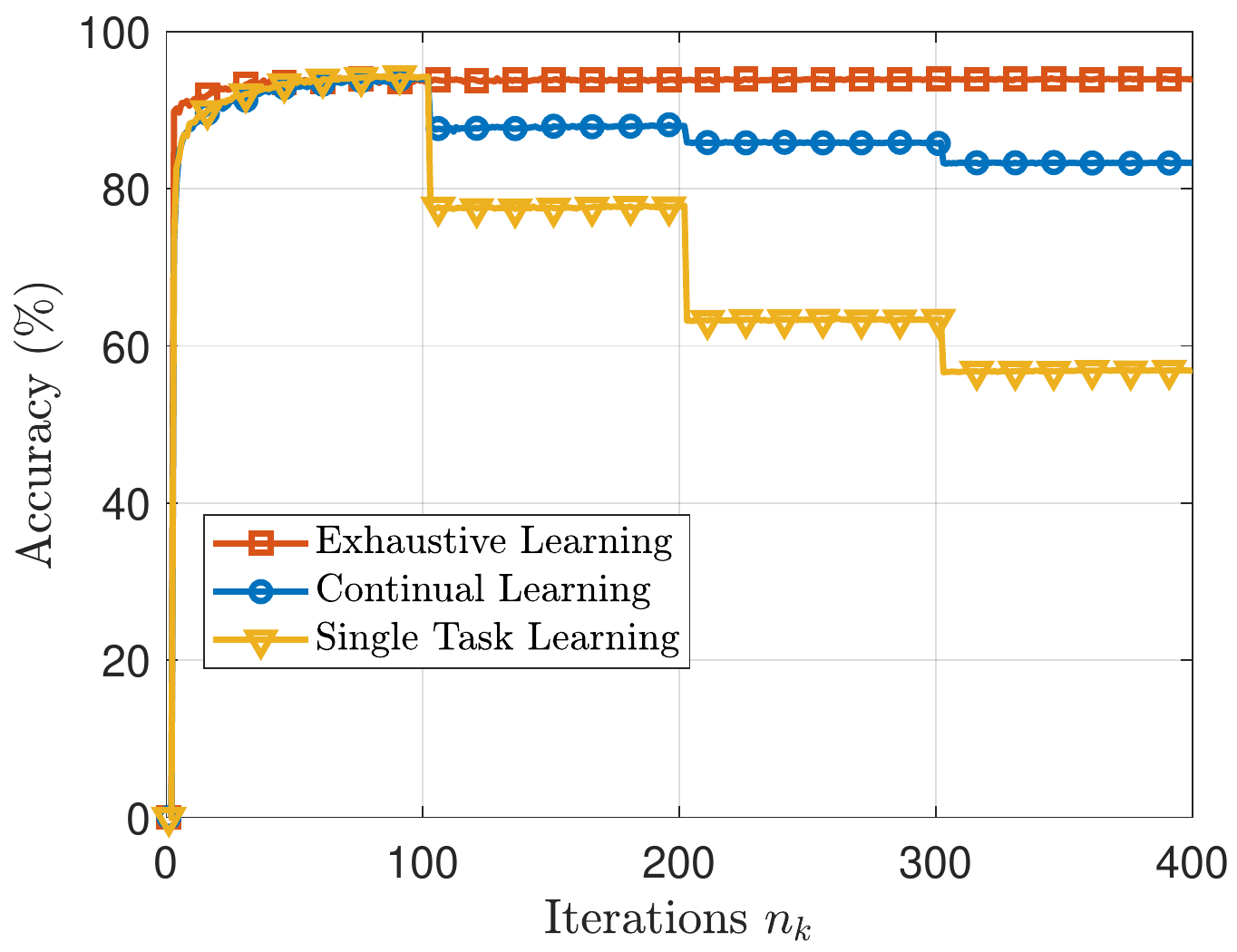}
	\caption{\small{Accuracy (\%) over first episode versus iterations $n_k$}}
	\label{forgetting}
		\vspace{-0.7cm}
\end{figure}
\indent Fig.~\ref{forgetting} compares the model robustness to \emph{catastrophic forgetting} upon training the different methods over $4$ sequential episodes. This is verified by measuring the model's accuracy on the first episode after training on each episode successively.  
Initially, it can be seen that all the methods achieve a similar accuracy of $94\%$ over the first episode. As the number of episodes increases, we can see that the proposed \ac{CL} is capable of reaching an accuracy of $84\%$ in contrast to a $58\%$ attained accuracy by single task learning. Clearly, unlike the exhaustive method, the \ac{CL} methodology is a lightweight scheme that permits accumulating the gained knowledge without the need to revisit data from preceding episodes. That said, the proposed approach achieves an accuracy that is only $14\%$ lower than that of exhaustive learning. Henceforth, this showcases the resilience of the proposed \ac{CL} method to \emph{catastrophic forgetting} despite its minimal de-synchronization time.\\
\indent Fig.~\ref{tradeoff} presents the relationship between the training loss and de-synchronization time, where the \ac{CL} solution proposed is implemented in a single episode for the different values of $n_k$. Fig.~\ref{tradeoff} demonstrates the presence of an explicit tradeoff between training loss and the de-synchronization time when solving for the optimal value of $n_k^*$ that is dependent on $\alpha$. 
By varying $\alpha$ for $0 \leq \alpha \leq 1$, $\Delta T$ can increase with the increase in $n_k$ reaching $\SI{47}{s}$ for 100 iterations performed. Consequently, by increasing $n_k$, the parameters converge to their optimal values $\boldsymbol{w}_k^{*}$, which minimizes the training loss accordingly. 

\vspace{-0.2cm}
\section{Conclusion}
In this paper, we have proposed a novel edge \ac{CL} framework to enable a robust synergy between a \ac{CT} and \ac{PT} in a dynamic environment. To guarantee an accurate and synchronous \ac{CT}, we have posed its model update process as a dual objective optimization problem. We have adopted an \ac{EWC} technique to overcome the increase in the de-synchronziation time. To address the stability-plasticity tradeoff accompanying the progressive growth of the regularization terms, we have proposed a modified \ac{CL} paradigm that considers a fair execution between historical episodes. Ultimately, we have shown that the proposed approach can achieve an accurate and synchronous CT that can balance between adaptation and stability.
\bibliographystyle{IEEEtran}
\def\baselinestretch{0.82}
\bibliography{bibliography}
\end{document}